\documentclass[11pt]{article}

\usepackage[preprint]{acl}

\usepackage{times}
\usepackage{latexsym}

\usepackage[T1]{fontenc}

\usepackage[utf8]{inputenc}

\usepackage{microtype}

\usepackage{inconsolata}

\usepackage{tabularx}
\usepackage{makecell}
\usepackage{graphicx}
\usepackage{algorithm}
\usepackage{algorithmic}
\usepackage{amssymb}
\usepackage{amsmath}
\usepackage{booktabs}
\usepackage{multirow}
\usepackage{arydshln}
\usepackage{multirow}
\usepackage{bbding}
\usepackage{times}
\usepackage{helvet}
\usepackage{courier}
\usepackage{xcolor}
\usepackage{newfloat}
\usepackage{listings}
\usepackage{hyperref}
\hypersetup{hidelinks}
\DeclareCaptionStyle{ruled}{labelfont=normalfont,labelsep=colon,strut=off} 
\floatstyle{ruled}
\newfloat{listing}{tb}{lst}{}
\floatname{listing}{Listing}
\title{LBLLM: Lightweight Binarization of Large Language Models via Three-Stage Distillation}

\author{
\textbf{Siqing Song\textsuperscript{1,2}},
 \textbf{Chuang Wang\textsuperscript{1,2}}\thanks{Corresponding author.},
 \textbf{Yong Lang\textsuperscript{3}},
 \textbf{Yi Yang\textsuperscript{3}},
 \textbf{Xu-Yao Zhang\textsuperscript{1,2}}
\\
  \textsuperscript{1}MAIS, Institute of Automation, Chinese Academy of Sciences, China
\\
 \textsuperscript{2}School of Artificial Intelligence, University of Chinese Academy of Sciences, China
\\
 \textsuperscript{3}Central Media Technology Institute, Huawei
}
\newcolumntype{Y}{>{\centering\arraybackslash}X}
\begin{document}
\maketitle
\begin{abstract}
Deploying large language models (LLMs) in resource-constrained environments is hindered by  heavy computational and memory requirements. We present LBLLM, a lightweight binarization framework that achieves effective W(1+1)A4 quantization through  a novel three-stage quantization strategy. The framework proceeds as follows: (1) initialize a high-quality quantized model via PTQ; (2)  quantize binarized weights, group-wise bitmaps, and quantization parameters through layer-wise distillation while keeping activations in full precision; and (3) training learnable activation quantization factors to dynamically quantize activations to 4 bits. This decoupled design mitigates interference between weight and activation quantization, yielding greater training stability and better inference accuracy.  LBLLM, trained only using 0.016B tokens with a single GPU, surpasses existing state-of-the-art binarization methods  on  W2A4 quantization settings across tasks of language modeling, commonsense QA, and language understanding. These results demonstrate that extreme low-bit quantization of LLMs can be both practical and highly effective without introducing any  extra high-precision channels nor rotational matrices commonly used in recent PTQ-based works, offering a promising path toward efficient LLM deployment on resource-limited situations. 
\end{abstract}

\section{Introduction}

\begin{figure*}[h!]
\centering
\includegraphics[width=1.0\textwidth]{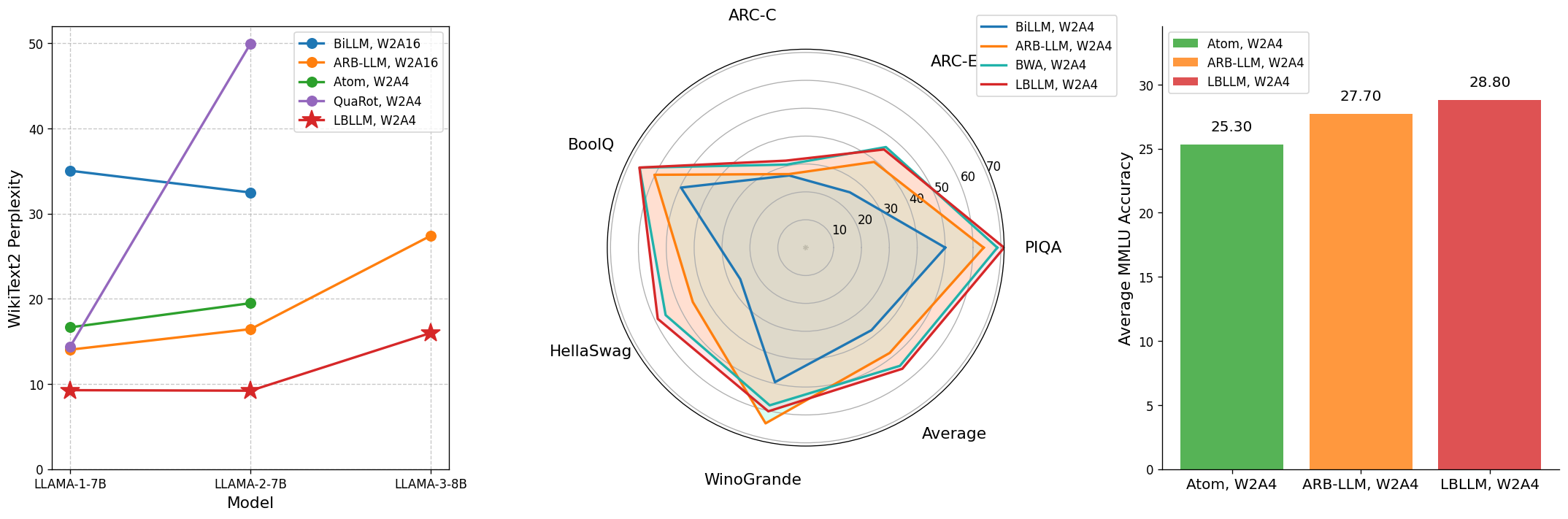} 
\caption{Left: Comparison of post-quantization perplexity between LBLLM and other methods across different LLAMA models on the W2A4 equivalent setting. Middle: Comparison of post-quantization commonsense QA performance on LLAMA-2-7B. Right: Comparison of average accuracy on the MMLU dataset after quantization across different LLAMA-1-7B models.}
\label{fig:results}
\end{figure*}

Large language models (LLMs) have recently gained significant attention, particularly for long-context reasoning tasks where both weights and activations dominate memory and compute costs \cite{10888431}. Quantization, which reduces precision by mapping high-bit parameters to low-bit representations, offers an effective means to jointly compress weights and activations and accelerate inference \cite{ouyang2024low, kurtic2025give}. However, most existing approaches still operate at relatively high bit-widths (typically $\geq$ 4 bits), limiting their potential for extreme compression in resource-constrained environments.


In joint weight and activation quantization, outlier activations, characterized by extreme magnitudes and long-tailed distributions (see Figure~\ref{fig:act_distribution}), pose a major challenge to maintaining accuracy. Moreover, extreme low-bit settings further constrain representational capacity, making sub-4-bit quantization seldom explored. Existing approaches address these issues in two ways: (1) auxiliary-channel methods (e.g., \cite{zhao2024atom}) retain high-precision channels to stabilize quantization, but these extra bits reduce compression efficiency and hinder inference speed due to mixed-precision execution; and (2) rotation-based methods (e.g., \cite{ashkboos2024quarot}) redistribute outliers via matrix transformations, which work well at W4A4 precision but degrade sharply at lower bit-widths and introduce additional computational and memory overhead.

Another line of research targets binarization, which significantly reduces bit-width but exacerbates quantization errors, often causing severe performance degradation. Existing binarization works typically quantize weights only, leaving activations in full precision (FP16). Depending on parameter optimization strategies, these methods fall into two categories: (1) Quantization-aware training (QAT) methods \cite{ma2024fbi, wang2024bitnet}, recover representational capability through extensive training but demand substantial computational resources, which requires dozens of 80GB GPUs and thousands of GPU-hours on datasets exceeding 10 billion tokens, limiting their practical use. (2) Post-training quantization (PTQ) methods \cite{huang2024billm,dong2024stbllm}, rapidly generate binarized models from pretrained checkpoints by employing an additional bit for fine-grained grouping and extra bits to encode outliers. However, these approaches frequently suffer from large accuracy degradation (more than 15 perplexity points on Wiki2) compared to full-precision models.

To enable efficient and high-quality binarization, we propose LBLLM, a lightweight framework that combines the strengths of PTQ and QAT. The model is first initialized via PTQ and then fine-tuned by a two-stage lightweight QAT that quantizes model weights and activation separately. Compared to prior PTQ-based methods, LBLLM improves perplexity by over 10 points and achieves accuracy comparable to full QAT approaches, while requiring only 0.016B tokens and a few dozen GPU hours on a single GPU to binarize a 7B model. Crucially, LBLLM avoids auxiliary high-precision channels and rotation matrices, enabling both effective memory compression and real inference acceleration.

We observed that directly extending existing lightweight layer-wise training \cite{ding2023cbq} to jointly quantize weights and activations to low bit-width  is ineffective. Quantization errors from weights and activations interfere during straight-through gradient estimation, resulting in unstable optimization. To address this issue, we decouple the QAT process into two stages: (1)  binarize weights and optimizing their quantization parameters while keeping activations in full precision; (2) subsequently quantize activations to 4 bits, introducing learnable clipping parameters to handle outliers and fine-tuning only quantization parameters. This staged approach achieves effective W(1+1)A4 quantization with minimal overhead.

The main contributions are summarized as follows.
\begin{itemize}

\item We introduce a three-stage quantization strategy combined PTQ initialization and lightweight QAT for joint weight-activation quantization, addressing the challenge of simultaneous low-bit quantization, which was not touched in previous layer-wise QAT training approaches.

\item Our hierarchical distillation strategy eliminates the need for high-precision auxiliary channels and rotation matrices commonly used in PTQ-based binarization approaches.

\item Extensive experiments demonstrate that our method surpasses state-of-the-art binarization on the W2A4 quantization level, while requiring only 0.016B tokens and a few dozen GPU hours on a single GPU.


\end{itemize}

\section{Related Works}

\begin{figure*}[t]
\centering
\includegraphics[width=1.0\textwidth]{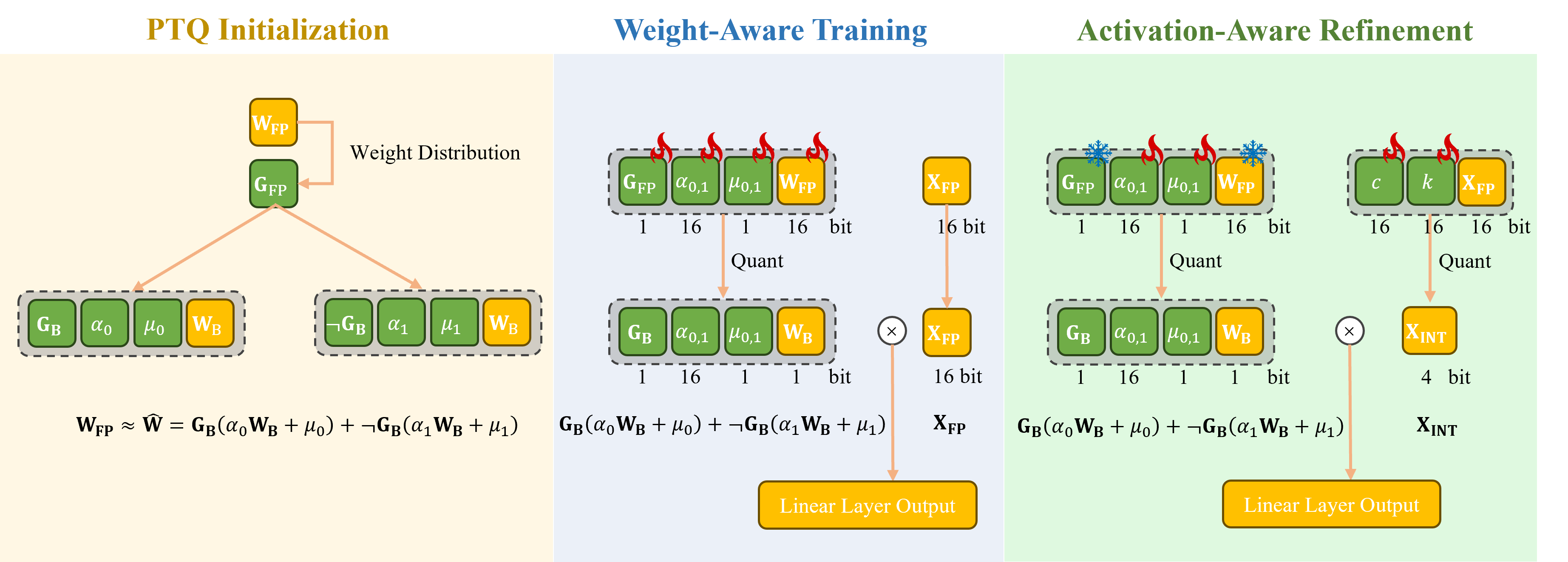} 
\caption{Illustration of the three-stage quantization strategy in LBLLM: Stage 1 uses a binarized PTQ method to obtain a high-quality initialized model; Stage 2 keeps activations in full precision and performs quantization-aware training on weights only; Stage 3 jointly quantizes activations and includes related parameters in training.}
\label{fig:method1}
\end{figure*}

\subsection{Binarization}

Binarization compresses both weights and activations into 1-bit representations \cite{park2025unifying}, achieving maximal storage savings and enabling highly efficient bitwise inference. However, applying binarization to LLMs is particularly challenging due to their structural complexity and the demanding nature of generative tasks \cite{gao2026vagu}.

Recent studies have proposed various strategies to address the challenges of LLM binarization. OneBit \cite{xu2024onebit} achieves weight-only binarization by introducing auxiliary full-precision parameters, while BitNet \cite{wang2023bitnet, wang2024bitnet} extends binary representations (${-1,+1}$) to ternary schemes (${-1,0,+1}$), effectively achieving 1.58-bit weights and low-bit activations. Although these methods improve training stability, they demand extensive computational resources, often requiring hundreds of high-end GPUs, which limits their practicality for broader applications. Moreover, most works focus on  weight binarization leaving the activations as full precision FP16 format.

To reduce computational cost, lighter PTQ-based methods such as BiLLM \cite{huang2024billm}, ARB-LLM \cite{liarb}, and BWA \cite{song2025achieving} employ bitmaps for fine-grained weight grouping and retain a small fraction of salient values in higher precision. While effective, these methods rely on extra bits for grouping and auxiliary high-precision channels to stabilize quantization, which limits compression efficiency and slows inference. Our work builds on the PTQ paradigm but introduces a lightweight QAT stage to reach W(1+1)A4 quantization on both weights and activations.

\subsection{Joint Weight and Activation Quantization}

Quantizing only model weights reduces storage cost \cite{liu2024llm, du2024bitdistiller}, but full acceleration with low-bit operators requires quantizing both weights and activations to lower bandwidth and computation, particularly for long-sequence inference \cite{jeon2025l4q, zhao2025ptq1, tao2025moqae, xiang2026fine}. The key challenge is handling activation outliers with extremely large values that cannot be accurately represented in low-bit formats.

Several methods address this by introducing auxiliary channels \cite{park2025outlier, su2025accurate}. LLM.int8 \cite{dettmers2022gpt3} preserves a small set of outlier channels in full precision, while Atom \cite{zhao2024atom} reorders channels based on Hessian values and quantizes outlier channels to 8 bits, using lower precision for the remaining values. Although effective, these approaches sacrifice compression efficiency and hardware friendliness due to their mixed-precision design.

An alternative line of work, including SmoothQuant \cite{xiao2023smoothquant} and AWQ \cite{lin2024awq}, balances quantization difficulty by applying equivalent rescaling transformations to weights and activations. Extensions such as QuaRot \cite{ashkboos2024quarot}, SpinQuant \cite{liu2024spinquant}, and DuQuant \cite{lin2024duquant} leverage Hadamard or learnable rotation matrices to achieve W4A4 quantization. However, these methods introduce additional storage and compute overhead, and their benefits diminish when targeting ultra-low-bit (2 bits) settings.

In contrast, our work removes the reliance on mixed-precision representations and rotation matrices, focusing  on  weight binarization with fine-grouping combined with 4-bit activation quantization, achieving higher compression  W(1+1)A4 and hardware efficiency without sacrificing accuracy.

\section{Preliminary}

This section provides the  background on LLM binarization, covering its scope,  weight binarization, fine-grained grouping, and a PTQ-based binarization scheme. 

\paragraph{Quantization in LLMs} Our work targets Transformer-based LLMs, focusing on linear layers and the KV cache, which together account for over 90\% of total computation. Less compute-intensive components, such as embeddings, normalization layers, and activation functions, remain in high precision. In each linear layer, both weights and input activations are quantized to the target bit-width prior to matrix multiplication, and the KV cache is quantized to the same bit-width as the activations. For example, W2A4 denotes 2-bit quantized weights (W2) and 4-bit quantized activations (A4).

\paragraph{Weight Binarization}
Weight binarization represents the extreme form of quantization, using a single bit to encode each weight. In linear layers, weights are approximated as:
\begin{equation}
\label{eq:binarization_our}
\textbf{W} \approx \textbf{W}_q = \alpha(\textbf{W}_B - \mu),
\end{equation}
where $\textbf{W}_B \in {0, 1}^{n \times m}$ is the binarized matrix, and $\alpha$, $\mu$ are full-precision (FP16) scaling and offset parameters. Here, $\alpha = \max(\textbf{W}) - \min(\textbf{W})$ and $\mu = -\lfloor \min(\textbf{W})/\alpha \rceil$; $\textbf{B}$ is stored using 1 bit per entry. In the QAT setting, $\alpha$ and $\mu$ are treated as learnable parameters and optimized jointly with other network weights, allowing the binarization scheme to better adapt to the data distribution.

\paragraph{Fine-Grained Weight Grouping}
Prior PTQ-based binarization \cite{huang2024billm} studies show that a single bit is often insufficient to preserve representational capacity. One strategy is to add an auxiliary bit as a bitmap to enable fine-grained element-wise grouping. Weights in the same group share quantization parameters, while different groups are quantized independently. This scheme effectively encodes four distinct states (equivalent to 2 bits) by combining one bit for grouping and one bit for the binary weight. Compared to a conventional 2-bit weight representation, it enables faster inference through efficient bitwise operations.

\paragraph{Activation Quantization and Binarization}
In long-sequence autoregressive LLM inference, activations and KV caches dominate memory and bandwidth usage, making their quantization essential. Following prior work, we quantize activations to 4 bits, represented either directly or as four binary channels, and jointly optimize them with binary weights and bitmaps for improved representational capacity. Previous PTQ methods, such as BWA \cite{song2025achieving}, search quantization parameters via clustering to achieve higher-quality binarization without requiring large datasets or high computation. 

 PTQ approaches still exhibit performance gaps relative to QAT-based methods. Moreover, PTQ methods often  rely on auxiliary channels to stabilize performance, leading to an average bit-width that falls short of the ideal compression target. In the subsequent section, we introduce LBLLM, which removes the additional channels using a lightweight training framework and achieves better performance.



\section{Method}

In this section, we introduce LBLLM, a lightweight binarization framework for LLMs based on QAT. We begin by outlining the overall three-stage quantization pipeline, followed by a description of the model adaptations used for quantization distillation,  including a relaxed formulation of binary weights and a progressively staged training strategy tailored for efficient optimization  using only a small fraction of the training data.

\subsection{Three-Stage Training}
As shown in Figure~\ref{fig:method1}, LBLLM employs a three-stage training pipeline. The process initializes the quantized model via a PTQ approach. In the second stage, we binarize the weights and distill weight-related parameters while keeping activation parameters in full precision. Finally, activation quantization layers are introduced, and a subset of both weight and activation quantization parameters is jointly fine-tuned.

\paragraph{Stage 1: PTQ Initialization}
We adopt the binarization method proposed in~\cite{song2025achieving} to obtain an initial quantized model, but discard its mixed-precision configuration in order to achieve higher compression efficiency. Specifically, we employ a fine-grained bitmap grouping structure and optimize the quantization parameters using a Hessian-weighted EM algorithm to minimize the quantization error of individual weight matrices. In table~\ref{tab:ablation_init},  experiments demonstrate  that a well-initialized PTQ model significantly improves the convergence of quantized models under lightweight training regimes.


\paragraph{Stage 2: Weight-Aware Training (WAT) }

We binarize the weights of all linear layers in the Transformer architecture, including those in both Attention and MLP modules, while keeping activations in full precision (FP16) during this stage. Training follows a layer-wise distillation strategy and employs the straight-through estimator (STE) to jointly update the quantization parameters $\mu$, $\alpha$, $\textbf{G}$, and the original weight matrix $\textbf{W}$ for each layer.



\paragraph{Stage 3: Activation-Aware Refinement (AAR) }

In the final stage, we fine-tune the model while dynamically quantizing activations. To address the prominent long-tail distribution characteristics of LLM activations, we propose a differentiable distribution-aware quantizer. This mechanism aims to adaptively partition the core dense region from sparse outlier regions using learnable knee points. It assigns distinct quantization step sizes to each region while employing learnable clipping factors to mitigate the impact of extreme outliers

\begin{equation}
\begin{split}
\label{eq:act_quantize}
    \hat{\textbf{X}} = & \sum_{j=1}^3 \mathbb{I}(k_{j-1} \le \textbf{X} < k_j) \\
    & \cdot \left( \left( \text{clamp}\left(\left\lfloor \frac{\textbf{X}}{\alpha_j} + \mu_j \right\rceil, 0, b_j\right) - \mu_j\right) \cdot \alpha_j \right), \\
    &\alpha = \frac{c_\alpha\text{max}(\textbf{X})-c_\beta\text{min}(\textbf{X})}{2^{k}-1}, \\
   &\mu = -\lfloor \frac{c_\beta\text{min}(\textbf{X})}{\alpha} \rceil,
\end{split}
\end{equation}

where $k_j$ denotes the trainable knee points. $c_\alpha$ and $c_\beta$ are trainable clipping factors. $\mathbb{I}(\cdot)$ is the hard indicator function. For the $j$-th region, $\alpha_j, \mu_j$ and $b_j$ represent the quantization parameters and assigned bit-width, respectively. In this stage, we update only the quantization parameters $(\mu, \alpha)$, while keeping the binarized weights $\textbf{W}$ and group assignments $\textbf{G}$ fixed. This refinement reduces activation quantization error and improves overall model accuracy as shown in Table~\ref{tab:ablation_train}.



\subsection{Lighter QAT for Binarization}

\begin{figure}[t]
\centering
\includegraphics[width=0.9\columnwidth]{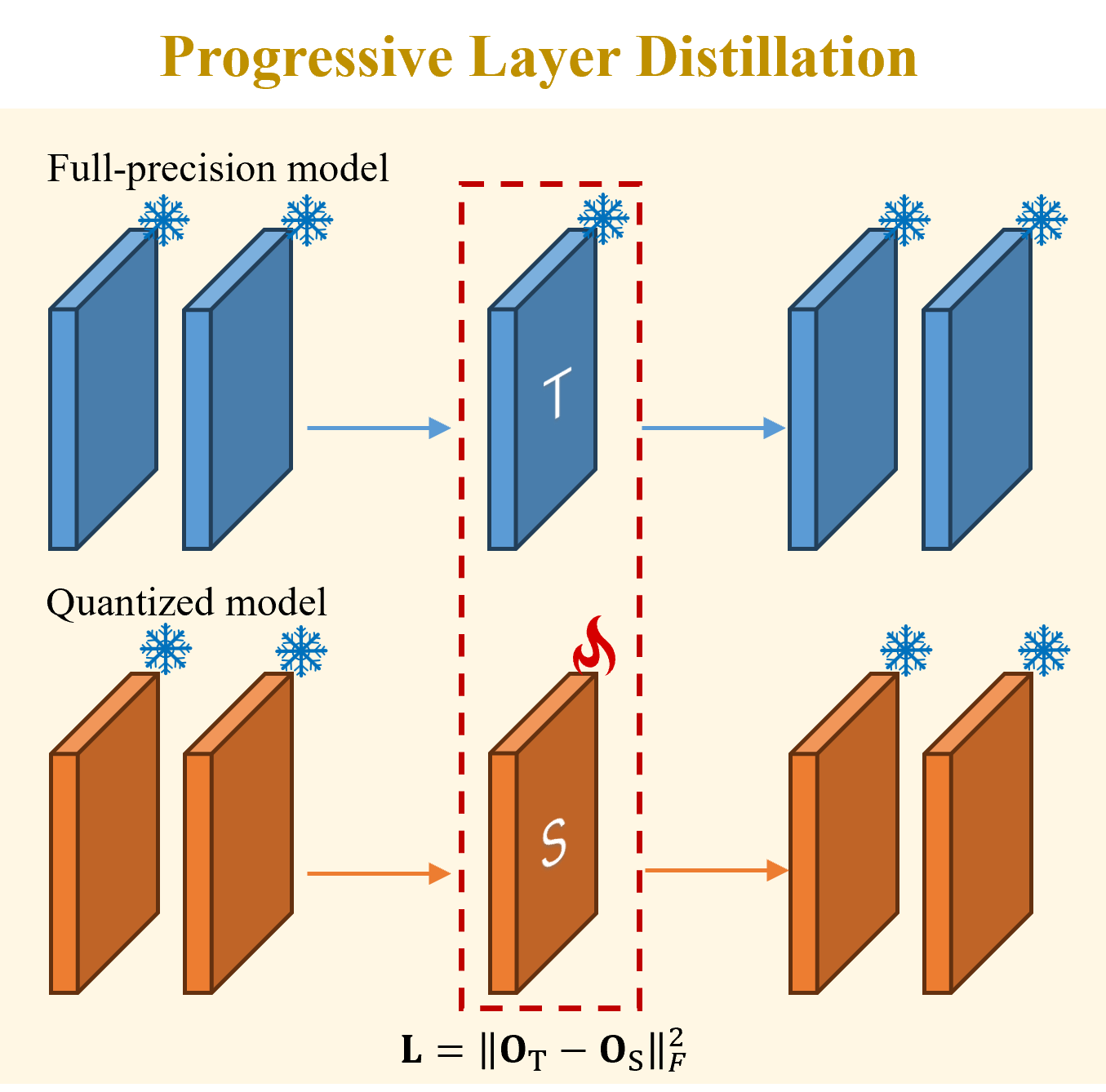} 
\caption{Illustration of the hierarchical distillation structure for LLMs, where each decoder layer is distilled progressively layer by layer.}
\label{fig:method2}
\end{figure}

LBLLM retains the same bianrized model architecture as prior PTQ approach \cite{liarb, song2025achieving}, but eliminates the auxiliary high-precision channels, resulting in a fully quantized W(1+1)A4 format. Unlike prior works  relying on architecture-specific tricks, such as rotation matrices \cite{ashkboos2024quarot} or extra high-precision channels, LBLLM operates without modifying the model structure.

\paragraph{Learning Group Bitmap}
The $W(1+1)$ weight parameterization consists of a binary weight matrix $\textbf{W}_B$, a binary group-assignment matrix $\textbf{G}_B$, and two pairs of shared scaling and offset parameters $(\alpha_g, \mu_g)$ for $g \in {0,1}$ corresponding to the two groups. 

To enable gradient-based optimization, the binary variables $\textbf{W}_B$ and $\textbf{G}_B$ are relaxed to full-precision counterparts $\textbf{W}_{FP}$ and $\textbf{G}_{FP}$ during training. A regularization term is applied to encourage these relaxed variables to gradually polarize toward binary values (0 or 1). In the forward pass, $\textbf{W}_{FP}$ and $\textbf{G}_{FP}$ are binarized via a clamp function to produce $\textbf{W}_B$ and $\textbf{G}_B$. For back-propagation, we employ the straight-through estimator (STE) to allow gradients to pass through the non-differentiable binarization step, ensuring stable and effective training of both weights and group assignments.

Specifically, the weight matrix is approximated  as
\begin{equation}
\label{eq:4}
   \textbf{W}_{\text{q}} = \textbf{G}_B(\alpha_0\textbf{W}_B+\mu_0)+\neg \textbf{G}_B(\alpha_1\textbf{W}_B+\mu_1),
\end{equation}
where $\textbf{G}_B$ denotes the group bitmap, and $\neg \textbf{G}_B$ is its negation, ensuring only one group is chosen for each element. During training, the relaxed group identifier $\textbf{G}_{FP}$ is encouraged to converge to either 0 or 1 via a regularization loss
\begin{equation}
\label{eq:5}
   \mathcal{L}_{reg} = \Vert \textbf{I} - \left| 2\textbf{G}_{FP} - \textbf{I} \right|^{\beta} \Vert_{F}^{2}, \, \textbf{I} \in 1^{n \times m},
\end{equation}
where $\beta$ is an annealing factor decreasing from 1 to 0. In the early training period, we encourage higher variability in the bitmap to explore a broader solution space. As training progresses, the value of $\beta$ is gradually reduced, driving the bitmap values polarized to 0 or 1.

\paragraph{Learning Knee Point}
 In AAR stage, we introduce a differentiable distribution-aware quantizer to improve activation quantization with learnable knee points and clipping factors. However, direct region partitioning based on hard threshold blocks gradient flow during backpropagation, preventing the optimization of knee point positions via gradient descent. To overcome this non-differentiability bottleneck, we introduce a temperature-scaled soft masking mechanism as

\begin{equation}
\label{eq:knee_backward}
    \tilde{\mathbb{I}}_j(\textbf{X}) = \text{sg}\left[ \mathbb{I}(\textbf{X} \in \mathcal{R}_j) - \pi_j(\textbf{X}; \tau) \right] + \pi_j(\textbf{X}; \tau).
\end{equation}

During training, $\tilde{\mathbb{I}}_j(\textbf{X})$ serves as the surrogate mask, where $\text{sg}[\cdot]$ denotes the stop-gradient operator, and $\pi_j(\textbf{X}; \tau)$ is the soft probability derived from the Sigmoid function with temperature $\tau$. We retain discrete hard partitioning during forward propagation to ensure quantization efficiency, while utilizing soft masks for gradient computation during backpropagation.

\paragraph{Progressive Layer Distillation}



As shown in Figure~\ref{fig:method2}, we formulate quantization as a layer-wise distillation process, treating the full-precision model as the teacher and the quantized model as the student \cite{shao2023omniquant, chen2024efficientqat}. Transformer-based LLMs are composed of sequentially stacked decoder layers with identical structures; we regard each decoder layer as the minimal training unit and optimize them progressively from shallow to deep. The input to each layer is the output of the previously quantized layer, allowing downstream layers to perceive and compensate for quantization errors accumulated earlier. The reconstruction loss is defined as
\begin{equation}
\label{eq:3}
   \mathcal{L}_{rec} = \Vert \textbf{O}_{\text{T}} - \textbf{O}_{\text{S}}\Vert_{F}^{2},
\end{equation}
where $\textbf{O}_{\text{T}}$ and $\textbf{O}_{\text{S}}$ denote the outputs of Decoderlayer in the teacher and student models respectively. This progressive distillation strategy reduces peak memory usage and enables fine-grained optimization of quantization parameters and binarized weights within each layer.

The layer-wise distillation loss is given by
$
   \mathcal{L}  = \mathcal{L}_{rec} + \lambda\mathcal{L}_{reg},$
where $\lambda$ is a hyperparameter used to balance the reconstruction loss and the regularization term.

\subsection{Discusion}
We compare LBLLM with earlier works on binarized PTQ and QAT approaches as well as the  layer-wise training strategy.


Prior PTQ-based binarization methods \cite{liarb, song2025achieving} demonstrate efficient weight binarization using bitmap-based grouping but rely on auxiliary high-precision channels and do not compress activations. In contrast, LBLLM adopts a similar fine-grained grouping strategy yet removes auxiliary channels entirely, achieving higher compression and extending binarization to activations. Compared to QAT-based approaches \cite{wang2024bitnet, ma2024fbi}, which achieve strong performance but require massive computational resources, LBLLM leverages PTQ-based initialization and a lightweight layer-wise QAT scheme, attaining comparable accuracy with drastically lower training cost.

Recent works have adopted layer-wise training for low-bit weight quantization \cite{chen2024efficientqat, ding2023cbq} but did not realize the challenges of jointly quantizing activations, majorly focusing on the setting of W4A16 or W2A16. Unlike weight-only quantization, simultaneous weight–activation quantization is considerably harder as quantization errors from both sources interact and can destabilize optimization. Our experiments in Table~\ref{tab:ablation_train} confirm that training a W(1+1)A4 model with a layer-wise strategy leads to unstable divergence.  LBLLM decouples weight and activation quantization as separated two stages, which mitigates error interference and stabilizes training.

\section{Experiments}

\begin{table*}[ht]
\renewcommand{\arraystretch}{1.3}
\begin{small}
\centering
\begin{tabularx}{\textwidth}{l:l:YYY:YYYYYYY}
\Xhline{1.2pt}
\textbf{Method} & \textbf{Bits} & \textbf{Wiki} & \textbf{PTB} & \textbf{C4} & \textbf{PIQA} & \textbf{ARCe} & \textbf{ARCc} & \textbf{BoolQ} & \textbf{Hella.} & \textbf{Wino.} & \textbf{Avg} \\
\hline
FP16 & -- & 5.47 & 22.51 & 6.97 & 76.93 & 53.58 & 40.53 & 71.07 & 72.96 & 67.17 & 63.71 \\
\hline
CBQ & W4A4 & 11.32 & -- & 12.56 & 68.25 & \textbf{46.23} & 31.56 & -- & 57.34 & -- & -- \\
\hline
BiLLM & W(1+1)A4 & 32.48 & 3877.38 & 40.52 & 50.05 & 25.38 & 26.54 & 49.63 & 26.05 & 49.49 & 37.86 \\
ARB-LLM & W(1+1)A4 & 19.14 & 165.46 & 20.88 & 63.82 & 39.31 & 27.05 & 60.21 & 44.94 & 54.54 & 48.31 \\
BWA & W(1+1)A4 & \textbf{8.89} & 69.46 & 12.74 & 68.72 & 46.13 & 30.55 & 66.12 & 55.76 & 58.01 & 54.22 \\
BWA w/o C & W(1+1)A4 & 243.73 & 5439.27 & 433.00 & -- & -- & -- & -- & -- & -- & -- \\
QuaRot & W2A4 & 49.98 & 571.22 & 80.14 & 54.41 & 28.45 & 23.21 & 57.89 & 28.57 & 48.15 & 40.11 \\
\hline
LBLLM & W(1+1)A4 & 9.18 & \textbf{36.79} & \textbf{10.91} & \textbf{70.97} & 44.65 & \textbf{32.51} & \textbf{64.50} & \textbf{58.54} & \textbf{58.27} & \textbf{54.91} \\
\Xhline{1.2pt}
\end{tabularx}
\caption{Perplexity and zero-shot accuracy results of different quantization methods on LLAMA-2-7B. For CBQ, since the code is not publicly available, we directly adopt the quantization settings and results reported in the original paper, while all other methods use a consistent W2A4 configuration. "BWA w/o C" denotes the performance of the BWA method after removing the additional channels. The best results are highlighted in bold.}
\label{tab:main_result}
\end{small}
\end{table*}

\subsection{Setup}

In this work, we quantize all linear layer weights in the LLM into 1-bit Boolean matrices along with 1-bit fine-grained group bitmaps. Activations are dynamically and asymmetrically quantized to 4 bits. The group size is 128. For the KV cache, we ensure that it is quantized to the same bit-width as the activations. No additional high-precision parameters are retained in our quantized model. To ensure a consistent experimental setup, all training is conducted on a single 80GB GPU.


\paragraph{Models and Datasets}
We evaluate our method on the open-source LLaMA-1 \cite{touvron2023llama}, LLaMA-2 \cite{touvron2023llama2}, and LLaMA-3 models. For training, we randomly sample 8,192 sequences (approximately 0.016B tokens) from the RedPajama corpus \cite{weber2024redpajama}, which contains over 100B tokens. This small-scale subset highlights the data efficiency of our approach while ensuring consistency across model sizes.

We evaluate performance across three categories: language generation, commonsense QA, and language understanding. For language generation, we report perplexity on WikiText-2 \cite{merity2016pointer}, PTB \cite{marcus1994penn}, and C4 \cite{raffel2020exploring}; for C4, we randomly sample 256 sequences of length 2048 from the test set, while for other datasets we use the full test split. Commonsense QA is assessed via zero-shot accuracy on PIQA \cite{bisk2020piqa}, ARC \cite{clark2018think}, BoolQ \cite{clark2019boolq}, HellaSwag \cite{zellers2019hellaswag}, and WinoGrande \cite{sakaguchi2021winogrande}. For language understanding, we evaluate on the MMLU benchmark \cite{hendrycks2020measuring}.

\paragraph{Baselines}

We compare LBLLM against existing weight-binarization methods such as BiLLM, ARB-LLM, and BWA \cite{huang2024billm, liarb, song2025achieving}; the weight and activation joint quantization method, QuaRot\cite{ashkboos2024quarot}, which employs rotation matrices; and CBQ\cite{ding2023cbq}, a lightweight training method based on progressive layer-wise distillation. As CBQ \cite{ding2023cbq} does not provide open-source code, we used the results reported in the original paper under the W4A4 setting for comparison. 

\subsection{Main Results}

\begin{table*}[t]
\renewcommand{\arraystretch}{1.3}
\begin{small}
\centering
\begin{tabularx}{\textwidth}{l:l:YYYYYYYYYY}
\Xhline{1.2pt}
\textbf{Model} & \textbf{Bits} & \textbf{Wiki} & \textbf{PTB} & \textbf{C4} & \textbf{PIQA} & \textbf{ARCe} & \textbf{ARCc} & \textbf{BoolQ} & \textbf{Hella.} & \textbf{Wino.} & \textbf{Avg} \\
\hline
\multirow{2}{*}{\textbf{LLAMA-1-7B}}
            & FP16 & 5.68 & 27.34 & 7.08 & 77.37 & 52.48 & 41.38 & 73.06 & 73.00 & 67.01 & 64.05\\
            \cline{2-12}
            & W(1+1)A4  & 9.08 & 37.46 & 10.42 & 71.33 & 44.74 & 33.53 & 66.24 & 58.29 & 61.72  & 55.97 \\
\hline
\multirow{2}{*}{\textbf{LLAMA-2-7B}}
            & FP16 & 5.47  & 22.51  & 6.97 & 76.93 & 53.58 & 40.53 & 71.07 & 72.96 & 67.17 & 63.71\\
            \cline{2-12}
            & W(1+1)A4   & 9.18 & 36.79 & 10.91 & 70.97 & 44.65 & 32.51 & 64.50 & 58.54 & 58.27 & 54.91 \\
\hline
\multirow{2}{*}{\textbf{LLAMA-3-8B}}
            & FP16 & 6.14 & 10.60 & 8.89 & 80.96 & 77.48 & 53.67 & 80.95 & 79.13 & 73.16  & 74.23  \\
            \cline{2-12}
            & W(1+1)A4  & 17.56 & 26.40 & 18.97 & 67.90 & 53.11 & 32.17 & 70.28 & 60.05 & 57.77 &  56.88 \\
\Xhline{1.2pt}
\end{tabularx}
\caption{Quantized perplexity and zero-shot accuracy results of LBLLM on different LLAMA models.}
\label{tab:diff_model_result}
\end{small}
\end{table*}


\paragraph{Language Generation Tasks}
We evaluate the perplexity of the quantized models on multiple datasets. Our training set, RedPajama, includes portions of WikiText2 and C4 but excludes PTB. Therefore, perplexity evaluation on PTB serves as an indicator of the quantized model’s overfitting tendency. 

Table~\ref{tab:main_result} presents the perplexity comparisons of various quantization methods on LLAMA-2-7B. It is evident that LBLLM achieves SOTA performance across all datasets and significantly outperforms a range of baseline methods. CBQ also employs layer-wise distillation training. Experiments demonstrate that our method achieves better performance in simultaneous weight and activation quantization. Specifically, LBLLM with W(1+1)A4 surpasses CBQ's results with W4A4. 

The perplexity of LBLLM is slightly higher than that of BWA only on the WikiText2 dataset. However, BWA exhibits signs of overfitting on WikiText2 and employs a mixed-precision representation, which implies additional channel overhead. Compared to the setting of BWA without any additional channels, which is under the exactly same compression ratio and model structure, LBLLM demonstrates significantly superior performance. 

Table~\ref{tab:diff_model_result} demonstrates the quantization performance of LBLLM on different versions of  LLAMA models, further validating the generalization capability of our quantization approach.

\paragraph{Zero-Shot Common Sense QA Tasks}
We  assess the effectiveness of our quantization method on several commonsense question answering tasks. Table~\ref{tab:main_result} provides a comparison of accuracy across different quantization methods on LLAMA-2-7B. Table~\ref{tab:diff_model_result} presents the quantization performance of LBLLM across different models. The results show that LBLLM surpasses previous SOTA methods and further approaches the performance of the full-precision model.

\paragraph{Language Understanding Tasks}
We evaluate the language understanding ability of the LBLLM-quantized model on the MMLU benchmark. Figure~\ref{fig:results} presents the performance of the quantized model based on LLAMA-1-7B across various sub-domains. Despite quantizing the full-precision model to an extremely low bit-width, it maintains a high level of language understanding performance.

\subsection{Ablation Study}
We conduct ablation studies on  on LLAMA-2-7B with a unified quantization setting of W(1+1)A4 to test different training strategy, settings of trainable parameters, and initialization methods.

\paragraph{Training Strategy}

\begin{table}[h]
\begin{small}
\begin{center}
\begin{tabular}{ccc:cccc}
\toprule
\textbf{WAT}& \textbf{AAR} & \textbf{DT} & \textbf{Wiki}$\downarrow$ & \textbf{PTB}$\downarrow$ & \textbf{C4}$\downarrow$ \\
\midrule
\XSolidBrush & \XSolidBrush & \XSolidBrush & 243.73 & 5439.27 & 433.00 \\
\XSolidBrush & \XSolidBrush & \Checkmark & NAN & NAN & NAN \\
\Checkmark & \XSolidBrush & \XSolidBrush & 9.40 & 42.86 & 11.08 \\
\Checkmark & \Checkmark & \XSolidBrush & \textbf{9.18} & \textbf{36.79} & \textbf{10.91} \\
\bottomrule
\end{tabular}
\caption{Ablation study of different training stages. DT refers to jointly quantizing weights and activations followed by direct layer-wise distillation. NAN indicates that the PPL computed from the model output is excessively large.}
\label{tab:ablation_train}
\end{center}
\end{small}
\end{table}

To verify the effectiveness of our decoupled training strategy for weight binarization and synchronized activation quantization, Table~\ref{tab:ablation_train} compares the average perplexity before and after each training stage. Experimental results show that the quantized model obtained after the fine-tuning on weights demonstrates a significant improvement over the initial quantized model. Although the activation-specific tuning requires low-bit quantization of activations, which leads to a performance drop compared to the first-stage result, the final performance after tuning confirms that our optimization mitigates the degradation caused by activation quantization errors.

\paragraph{Trainable Parameters}

\begin{table}[h]
\renewcommand{\arraystretch}{1.5}
\begin{small}
\centering
\begin{tabularx}{\linewidth}{YYY:YYYY}
\Xhline{1.2pt}
\textbf{$\alpha,\mu$} & \textbf{G} & \textbf{W} & \textbf{Wiki}$\downarrow$ & \textbf{PTB}$\downarrow$ & \textbf{C4}$\downarrow$ & \textbf{Avg.}$\downarrow$\\
\hline
\Checkmark & \XSolidBrush & \XSolidBrush & 18.61 & 387.99 & 17.99 & 141.53 \\
\XSolidBrush & \XSolidBrush & \Checkmark & 15.16 & 1638.36 & 15.97 & 556.50 \\
\Checkmark & \Checkmark & \Checkmark & \textbf{9.48} & \textbf{69.26} & \textbf{11.20} & \textbf{29.98} \\
\Xhline{1.2pt}
\end{tabularx}
\caption{Ablation study on the impact of parameter participation in training each component of WAT on the final performance, all experimental results are conducted on LLAMA-2-7B under the W(1+1)A16 quantization setting.}
\label{tab:ablation_para_1}
\end{small}
\end{table}

\begin{table}[h]
\begin{small}
\begin{center}
\begin{tabular}{cc:cccc}
\toprule
\textbf{$\alpha,\mu$}& \textbf{$c,k$} & \textbf{Wiki}$\downarrow$ & \textbf{PTB}$\downarrow$ & \textbf{C4}$\downarrow$ & \textbf{Avg. PPL}$\downarrow$\\
\midrule
\XSolidBrush & \XSolidBrush & 9.40 & 42.86 & 11.08 & 21.12 \\
\Checkmark & \XSolidBrush & 9.51 & 40.31 & 10.99 & 20.27 \\
\XSolidBrush & \Checkmark & 9.39 & 44.08 & 11.09 & 21.52 \\
\Checkmark & \Checkmark & \textbf{9.18} & \textbf{36.79} & \textbf{10.91} & \textbf{18.96} \\
\bottomrule
\end{tabular}
\caption{Ablation study on the impact of parameter participation in training each component of AAR on the final performance.}
\label{tab:ablation_para_2}
\end{center}
\end{small}
\end{table}

To examine the influence of different parameter training strategies on the final model performance, we compare the impact of parameter inclusion/exclusion on the quantized model performance in Table~\ref{tab:ablation_para_1} and Table~\ref{tab:ablation_para_2}. The experimental results confirm that global optimization across all parameters contributes to achieving a higher-quality quantized model.

\paragraph{Initialization Method}

\begin{table}[h]
\begin{small}
\begin{center}
\begin{tabular}{c:cccc}
\toprule
\textbf{Initialization} & \textbf{Wiki}$\downarrow$ & \textbf{PTB}$\downarrow$ & \textbf{C4}$\downarrow$ & \textbf{Avg. PPL}$\downarrow$\\
\midrule
RTN & NAN & NAN & NAN & NAN \\
BWA & \textbf{9.48} & \textbf{69.26} & \textbf{11.20} & \textbf{29.98} \\
\bottomrule
\end{tabular}
\caption{Ablation study results using different initialization methods. NAN indicates that the PPL computed from the model output is excessively large.}
\label{tab:ablation_init}
\end{center}
\end{small}
\end{table}

Table~\ref{tab:ablation_init} compares the effect of initializing the quantized model with or without PTQ-based methods. By contrasting the performance of models initialized via RTN quantization and subsequently fine-tuned, we demonstrate that better initialization schemes can improve model performance under our lightweight framework.

\section{Limitations}
Although LBLLM takes a significant step toward practical and efficient LLM deployment in resource-constrained environments, challenges remain in removing fine-grained grouping and further narrowing the gap with full-precision models. Empirically, the primary bottleneck in the W(1+1) configuration lies in the joint optimization of the 1-bit weight representation and the 1-bit grouping parameter. This process currently relies heavily on PTQ-derived priors for initialization, an area we intend to investigate further in future work. Moreover, current inference efficiency gains are predominantly software-based and yield marginal returns, often necessitating bespoke CUDA kernel implementations for different quantization schemes. True computational breakthroughs require dedicated hardware support, which remains largely in the research phase; we hope subsequent studies will bridge this gap.

\section*{Acknowledgments}
This work is supported by National Key R\&D Program of China (No. 2025ZD0122000), the National Natural Science Foundation of China (No. 62576343), and  the Strategic Priority Research Program of the Chinese Academy of Sciences (No.XDA0480200). Chuang Wang, Siqing Song, and Xuyao Zhang are also supported by the funding from Central Media Technology Institute, Huawei.  Codes would be available at \url{https://github.com/JimmyCrave/LBLLM} upon internal screen approval.



\bibliography{custom}

@inproceedings{huang2024billm,
  title={BiLLM: pushing the limit of post-training quantization for LLMs},
  author={Huang, Wei and Liu, Yangdong and Qin, Haotong and Li, Ying and Zhang, Shiming and Liu, Xianglong and Magno, Michele and Qi, Xiaojuan},
  booktitle={Proceedings of the 41st International Conference on Machine Learning},
  pages={20023--20042},
  year={2024}
}

@article{dettmers2022gpt3,
  title={Gpt3. int8 (): 8-bit matrix multiplication for transformers at scale},
  author={Dettmers, Tim and Lewis, Mike and Belkada, Younes and Zettlemoyer, Luke},
  journal={Advances in Neural Information Processing Systems},
  volume={35},
  pages={30318--30332},
  year={2022}
}

@article{wang2023bitnet,
  title={Bitnet: Scaling 1-bit transformers for large language models},
  author={Wang, Hongyu and Ma, Shuming and Dong, Li and Huang, Shaohan and Wang, Huaijie and Ma, Lingxiao and Yang, Fan and Wang, Ruiping and Wu, Yi and Wei, Furu},
  journal={arXiv preprint arXiv:2310.11453},
  year={2023}
}

@article{wang2024bitnet,
  title={BitNet a4. 8: 4-bit Activations for 1-bit LLMs},
  author={Wang, Hongyu and Ma, Shuming and Wei, Furu},
  journal={arXiv preprint arXiv:2411.04965},
  year={2024}
}

@inproceedings{dong2024stbllm,
  title={STBLLM: Breaking the 1-Bit Barrier with Structured Binary LLMs},
  author={Dong, Peijie and Li, Lujun and Zhong, Yuedong and Du, Dayou and Fan, Ruibo and Chen, Yuhan and Tang, Zhenheng and Wang, Qiang and Xue, Wei and Guo, Yike and others},
  booktitle={Proceedings of the 38th International Conference on Neural Information Processing Systems},
  year={2024}
}

@inproceedings{liu2024spinquant,
  title={SpinQuant--LLM quantization with learned rotations},
  author={Liu, Zechun and Zhao, Changsheng and Fedorov, Igor and Soran, Bilge and Choudhary, Dhruv and Krishnamoorthi, Raghuraman and Chandra, Vikas and Tian, Yuandong and Blankevoort, Tijmen},
  booktitle={Proceedings of the 38th International Conference on Neural Information Processing Systems},
  year={2024}
}

@inproceedings{ashkboos2024quarot,
  title={QuaRot: outlier-free 4-bit inference in rotated LLMs},
  author={Ashkboos, Saleh and Mohtashami, Amirkeivan and Croci, Maximilian L and Li, Bo and Cameron, Pashmina and Jaggi, Martin and Alistarh, Dan and Hoefler, Torsten and Hensman, James},
  booktitle={Proceedings of the 38th International Conference on Neural Information Processing Systems},
  pages={100213--100240},
  year={2024}
}

@article{zhao2024atom,
  title={Atom: Low-bit quantization for efficient and accurate llm serving},
  author={Zhao, Yilong and Lin, Chien-Yu and Zhu, Kan and Ye, Zihao and Chen, Lequn and Zheng, Size and Ceze, Luis and Krishnamurthy, Arvind and Chen, Tianqi and Kasikci, Baris},
  journal={Proceedings of Machine Learning and Systems},
  volume={6},
  pages={196--209},
  year={2024}
}

@inproceedings{lin2024duquant,
  title={Duquant: Distributing outliers via dual transformation makes stronger quantized llms},
  author={Lin, Haokun and Xu, Haobo and Wu, Yichen and Cui, Jingzhi and Zhang, Yingtao and Mou, Linzhan and Song, Linqi and Sun, Zhenan and Wei, Ying},
  booktitle={The Thirty-eighth Annual Conference on Neural Information Processing Systems},
  year={2024}
}

@article{ma2024fbi,
  title={Fbi-llm: Scaling up fully binarized llms from scratch via autoregressive distillation},
  author={Ma, Liqun and Sun, Mingjie and Shen, Zhiqiang},
  journal={arXiv preprint arXiv:2407.07093},
  year={2024}
}

@article{lin2024awq,
  title={AWQ: Activation-aware Weight Quantization for On-Device LLM Compression and Acceleration},
  author={Lin, Ji and Tang, Jiaming and Tang, Haotian and Yang, Shang and Chen, Wei-Ming and Wang, Wei-Chen and Xiao, Guangxuan and Dang, Xingyu and Gan, Chuang and Han, Song},
  journal={Proceedings of Machine Learning and Systems},
  pages={87--100},
  year={2024}
}

@inproceedings{liu2024llm,
  title={Llm-qat: Data-free quantization aware training for large language models},
  author={Liu, Zechun and Oguz, Barlas and Zhao, Changsheng and Chang, Ernie and Stock, Pierre and Mehdad, Yashar and Shi, Yangyang and Krishnamoorthi, Raghuraman and Chandra, Vikas},
  booktitle={Findings of the Association for Computational Linguistics: ACL 2024},
  pages={467--484},
  year={2024}
}

@article{xu2024onebit,
  title={Onebit: Towards extremely low-bit large language models},
  author={Xu, Yuzhuang and Han, Xu and Yang, Zonghan and Wang, Shuo and Zhu, Qingfu and Liu, Zhiyuan and Liu, Weidong and Che, Wanxiang},
  journal={Advances in Neural Information Processing Systems},
  volume={37},
  pages={66357--66382},
  year={2024}
}

@article{touvron2023llama,
  title={Llama: Open and efficient foundation language models},
  author={Touvron, Hugo and Lavril, Thibaut and Izacard, Gautier and Martinet, Xavier and Lachaux, Marie-Anne and Lacroix, Timoth{\'e}e and Rozi{\`e}re, Baptiste and Goyal, Naman and Hambro, Eric and Azhar, Faisal and others},
  journal={arXiv preprint arXiv:2302.13971},
  year={2023}
}

@article{touvron2023llama2,
  title={Llama 2: Open foundation and fine-tuned chat models},
  author={Touvron, Hugo and Martin, Louis and Stone, Kevin and Albert, Peter and Almahairi, Amjad and Babaei, Yasmine and Bashlykov, Nikolay and Batra, Soumya and Bhargava, Prajjwal and Bhosale, Shruti and others},
  journal={arXiv preprint arXiv:2307.09288},
  year={2023}
}

@inproceedings{merity2016pointer,
  title={Pointer sentinel mixture models},
  author={Merity, Stephen and Xiong, Caiming and Bradbury, James and Socher, Richard},
  booktitle={Proceedings of the 30th International Conference on Neural Information Processing Systems},
  year={2016}
}

@inproceedings{marcus1994penn,
  title={The penn treebank: Annotating predicate argument structure},
  author={Marcus, Mitch and Kim, Grace and Marcinkiewicz, Mary Ann and MacIntyre, Robert and Bies, Ann and Ferguson, Mark and Katz, Karen and Schasberger, Britta},
  booktitle={Human Language Technology: Proceedings of a Workshop held at Plainsboro, New Jersey, March 8-11, 1994},
  year={1994}
}

@article{raffel2020exploring,
  title={Exploring the limits of transfer learning with a unified text-to-text transformer},
  author={Raffel, Colin and Shazeer, Noam and Roberts, Adam and Lee, Katherine and Narang, Sharan and Matena, Michael and Zhou, Yanqi and Li, Wei and Liu, Peter J},
  journal={Journal of machine learning research},
  volume={21},
  number={140},
  pages={1--67},
  year={2020}
}

@inproceedings{bisk2020piqa,
  title={Piqa: Reasoning about physical commonsense in natural language},
  author={Bisk, Yonatan and Zellers, Rowan and Gao, Jianfeng and Choi, Yejin and others},
  booktitle={Proceedings of the AAAI conference on artificial intelligence},
  volume={34},
  number={05},
  pages={7432--7439},
  year={2020}
}

@inproceedings{clark2019boolq,
  title={BoolQ: Exploring the Surprising Difficulty of Natural Yes/No Questions},
  author={Clark, Christopher and Lee, Kenton and Chang, Ming-Wei and Kwiatkowski, Tom and Collins, Michael and Toutanova, Kristina},
  booktitle={Proceedings of NAACL-HLT},
  pages={2924--2936},
  year={2019}
}

@article{sakaguchi2021winogrande,
  title={Winogrande: An adversarial winograd schema challenge at scale},
  author={Sakaguchi, Keisuke and Bras, Ronan Le and Bhagavatula, Chandra and Choi, Yejin},
  journal={Communications of the ACM},
  volume={64},
  number={9},
  pages={99--106},
  year={2021},
  publisher={ACM New York, NY, USA}
}

@article{zellers2019hellaswag,
  title={Hellaswag: Can a machine really finish your sentence?},
  author={Zellers, Rowan and Holtzman, Ari and Bisk, Yonatan and Farhadi, Ali and Choi, Yejin},
  journal={arXiv preprint arXiv:1905.07830},
  year={2019}
}

@article{clark2018think,
  title={Think you have solved question answering? try arc, the ai2 reasoning challenge},
  author={Clark, Peter and Cowhey, Isaac and Etzioni, Oren and Khot, Tushar and Sabharwal, Ashish and Schoenick, Carissa and Tafjord, Oyvind},
  journal={arXiv preprint arXiv:1803.05457},
  year={2018}
}

@inproceedings{xiao2023smoothquant,
  title={Smoothquant: Accurate and efficient post-training quantization for large language models},
  author={Xiao, Guangxuan and Lin, Ji and Seznec, Mickael and Wu, Hao and Demouth, Julien and Han, Song},
  booktitle={International Conference on Machine Learning},
  pages={38087--38099},
  year={2023},
  organization={PMLR}
}

@article{hendrycks2020measuring,
  title={Measuring massive multitask language understanding},
  author={Hendrycks, Dan and Burns, Collin and Basart, Steven and Zou, Andy and Mazeika, Mantas and Song, Dawn and Steinhardt, Jacob},
  journal={arXiv preprint arXiv:2009.03300},
  year={2020}
}

@inproceedings{ding2023cbq,
  title={Cbq: Cross-block quantization for large language models},
  author={Ding, Xin and Liu, Xiaoyu and Tu, Zhijun and Zhang, Yun and Li, Wei and Hu, Jie and Chen, Hanting and Tang, Yehui and Xiong, Zhiwei and Yin, Baoqun and others},
  booktitle={Proceedings of the 38th International Conference on Neural Information Processing Systems},
  year={2024}
}

@inproceedings{du2024bitdistiller,
  title={Bitdistiller: Unleashing the potential of sub-4-bit llms via self-distillation},
  author={Du, Dayou and Zhang, Yijia and Cao, Shijie and Guo, Jiaqi and Cao, Ting and Chu, Xiaowen and Xu, Ningyi},
  booktitle={Proceedings of the 62nd Annual Meeting of the Association for Computational Linguistics (Volume 1: Long Papers)},
  pages={102--116},
  year={2024}
}

@inproceedings{chen2024efficientqat,
    title = "{E}fficient{QAT}: Efficient Quantization-Aware Training for Large Language Models",
    author = "Chen, Mengzhao  and
      Shao, Wenqi  and
      Xu, Peng  and
      Wang, Jiahao  and
      Gao, Peng  and
      Zhang, Kaipeng  and
      Luo, Ping",
    booktitle = "Proceedings of the 63rd Annual Meeting of the Association for Computational Linguistics",
    year = "2025",
    pages = "10081--10100",
}

@inproceedings{shao2023omniquant,
  title={Omniquant: Omnidirectionally calibrated quantization for large language models},
  author={Shao, Wenqi and Chen, Mengzhao and Zhang, Zhaoyang and Xu, Peng and Zhao, Lirui and Li, Zhiqian and Zhang, Kaipeng and Gao, Peng and Qiao, Yu and Luo, Ping},
  booktitle={Proceedings of the 30th International Conference on Neural Information Processing Systems},
  year={2016}
}

@inproceedings{song2025achieving,
  title={Achieving binary weight and activation for LLMs using Post-Training Quantization},
  author={Song, Siqing and Wang, Chuang and Wang, Rui-Qi and Yang, Yi and Zhang, Xu-Yao},
  booktitle={Findings of the Association for Computational Linguistics: ACL 2025},
  pages={8782--8795},
  year={2025}
}

@article{weber2024redpajama,
	title   = {RedPajama: an Open Dataset for Training Large Language Models},
	author  = {Maurice Weber and Daniel Y. Fu and Quentin Anthony and Yonatan Oren and Shane Adams and Anton Alexandrov and Xiaozhong Lyu and Huu Nguyen and Xiaozhe Yao and Virginia Adams and Ben Athiwaratkun and Rahul Chalamala and Kezhen Chen and Max Ryabinin and Tri Dao and Percy Liang and Christopher Ré and Irina Rish and Ce Zhang},
	journal = {NeurIPS Datasets and Benchmarks Track},
	year    = 2024,
}

@inproceedings{liarb,
  title={Arb-llm: Alternating refined binarizations for large language models},
  author={Li, Zhiteng and Yan, Xianglong and Zhang, Tianao and Qin, Haotong and Xie, Dong and Tian, Jiang and Kong, Linghe and Zhang, Yulun and Yang, Xiaokang and others},
  booktitle={Proceedings of the 38th International Conference on Neural Information Processing Systems},
  year={2024}
}

@inproceedings{jeon2025l4q,
  title={L4Q: parameter efficient quantization-aware fine-tuning on large language models},
  author={Jeon, Hyesung and Kim, Yulhwa and Kim, Jae-Joon},
  booktitle={Proceedings of the 63rd Annual Meeting of the Association for Computational Linguistics (Volume 1: Long Papers)},
  pages={2002--2024},
  year={2025}
}

@article{zhao2025ptq1,
  title={Ptq1. 61: Push the real limit of extremely low-bit post-training quantization methods for large language models},
  author={Zhao, Jiaqi and Zhang, Miao and Wang, Ming and Shang, Yuzhang and Zhang, Kaihao and Guan, Weili and Wang, Yaowei and Zhang, Min},
  journal={arXiv preprint arXiv:2502.13179},
  year={2025}
}

@article{tao2025moqae,
  title={MoQAE: Mixed-Precision Quantization for Long-Context LLM Inference via Mixture of Quantization-Aware Experts},
  author={Tao, Wei and Lu, Haocheng and Qu, Xiaoyang and Zhang, Bin and Lu, Kai and Wan, Jiguang and Wang, Jianzong},
  journal={arXiv preprint arXiv:2506.07533},
  year={2025}
}

@article{park2025outlier,
  title={Outlier-Safe Pre-Training for Robust 4-Bit Quantization of Large Language Models},
  author={Park, Jungwoo and Lee, Taewhoo and Yoon, Chanwoong and Hwang, Hyeon and Kang, Jaewoo},
  journal={arXiv preprint arXiv:2506.19697},
  year={2025}
}

@article{su2025accurate,
  title={Accurate kv cache quantization with outlier tokens tracing},
  author={Su, Yi and Zhou, Yuechi and Qiu, Quantong and Li, Juntao and Xia, Qingrong and Li, Ping and Duan, Xinyu and Wang, Zhefeng and Zhang, Min},
  journal={arXiv preprint arXiv:2505.10938},
  year={2025}
}

@inproceedings{kurtic2025give,
  title={“Give Me BF16 or Give Me Death”? Accuracy-Performance Trade-Offs in LLM Quantization},
  author={Kurtic, Eldar and Marques, Alexandre Noll and Pandit, Shubhra and Kurtz, Mark and Alistarh, Dan},
  booktitle={Proceedings of the 63rd Annual Meeting of the Association for Computational Linguistics (Volume 1: Long Papers)},
  pages={26872--26886},
  year={2025}
}

@article{park2025unifying,
  title={Unifying Uniform and Binary-coding Quantization for Accurate Compression of Large Language Models},
  author={Park, Seungcheol and Bae, Jeongin and Kwon, Beomseok and Kim, Minjun and Kim, Byeongwook and Kwon, Se Jung and Kang, U and Lee, Dongsoo},
  journal={arXiv preprint arXiv:2506.03781},
  year={2025}
}

@article{ouyang2024low,
  title={Low-bit quantization favors undertrained llms: Scaling laws for quantized llms with 100t training tokens},
  author={Ouyang, Xu and Ge, Tao and Hartvigsen, Thomas and Zhang, Zhisong and Mi, Haitao and Yu, Dong},
  journal={arXiv preprint arXiv:2411.17691},
  year={2024}
}

@article{xiang2026fine,
  title={Fine-Grained Post-Training Quantization for Large Vision Language Models with Quantization-Aware Integrated Gradients},
  author={Xiang, Ziwei and Zeng, Fanhu and Fang, Hongjian and Wang, Rui-Qi and Chen, Renxing and Zhu, Yanan and Chen, Yi and Yang, Peipei and Zhang, Xu-Yao},
  journal={arXiv preprint arXiv:2603.17809},
  year={2026}
}

@INPROCEEDINGS{10888431,
  author={Gao, Shibo and Yang, Peipei and Huang, Linlin},
  booktitle={ICASSP 2025 - 2025 IEEE International Conference on Acoustics, Speech and Signal Processing (ICASSP)}, 
  title={SUVAD: Semantic Understanding Based Video Anomaly Detection Using MLLM}, 
  year={2025},
  volume={},
  number={},
  pages={1-5},
  doi={10.1109/ICASSP49660.2025.10888431}}

@inproceedings{gao2026vagu,
  title={Vagu \& gts: Llm-based benchmark and framework for joint video anomaly grounding and understanding},
  author={Gao, Shibo and Yang, Peipei and Liu, Yangyang and Chen, Yi and Zhu, Han and Zhang, Xu-Yao and Huang, Linlin},
  booktitle={Proceedings of the AAAI Conference on Artificial Intelligence},
  volume={40},
  number={6},
  pages={4167--4175},
  year={2026}
}

\appendix


\section{Difficulties in Joint Weight and Activation Quantization}

\paragraph{Activation Distribution}

\begin{figure}[h]
\centering
\includegraphics[width=1.0\columnwidth]{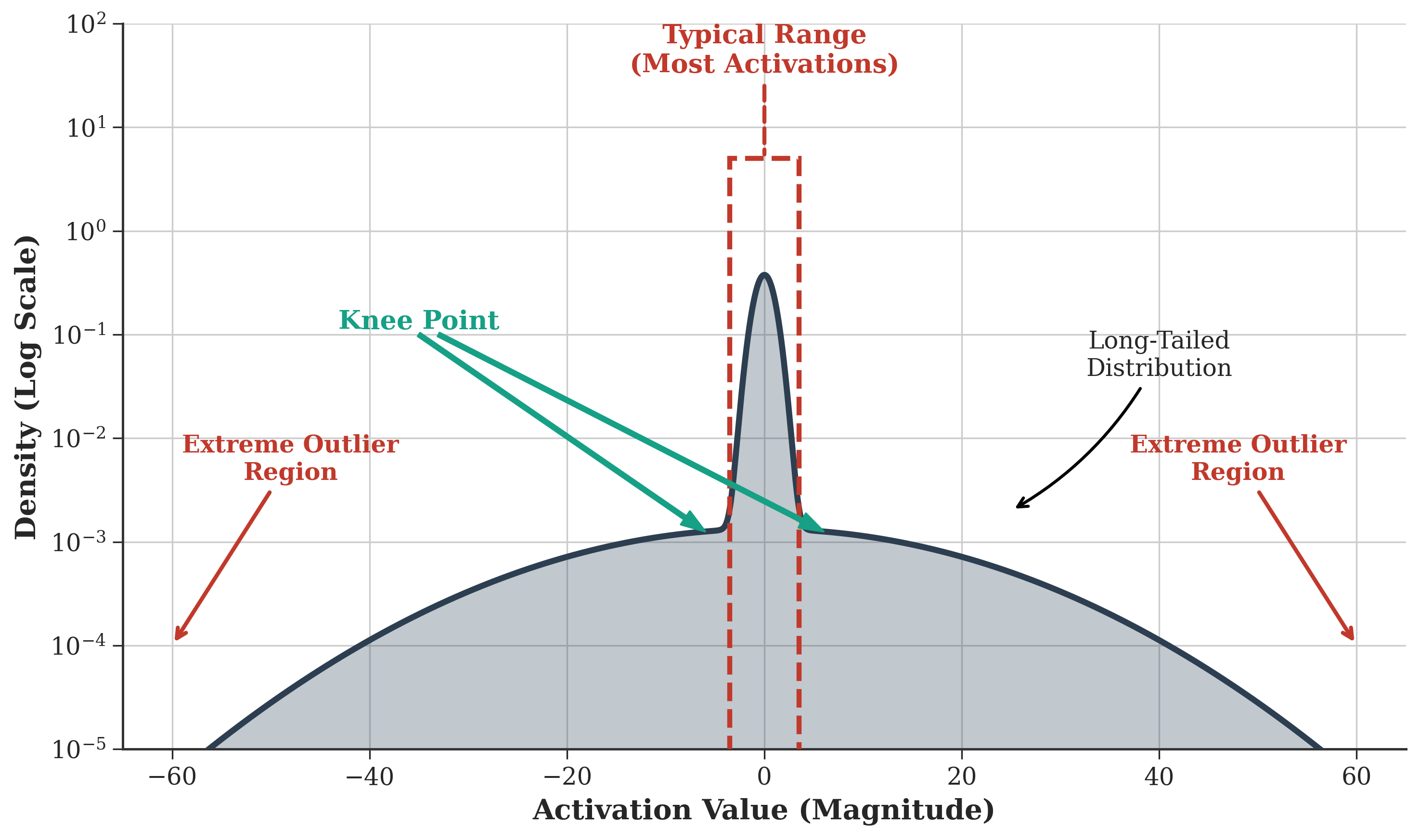} 
\caption{Illustration of activation distribution. The "Knee Point" marks the critical transition boundary where the density shifts significantly, separating the primary bell-shaped distribution from the sparse but magnitude-heavy outliers.}
\label{fig:act_distribution}
\end{figure}

As illustrated in Figure~\ref{fig:act_distribution}, activation values exhibit a pronounced long-tailed distribution. By employing a logarithmic scale, we visualize the distinct 'Knee Point,' which demarcates the typical range from the outlier regions. While the probability density of these outliers is minimal, their extreme magnitudes significantly expand the dynamic range, thereby degrading the quantization resolution for the majority of activations.

\paragraph{Error Coupling between Weights and Activations} 

\begin{figure}[h]
\centering
\includegraphics[width=1.0\columnwidth]{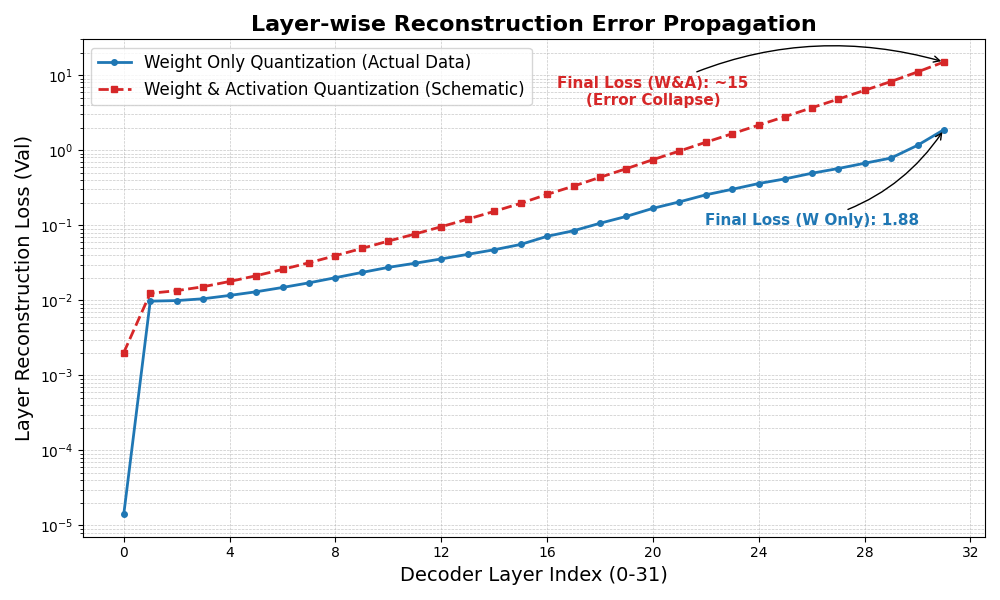} 
\caption{Layer-wise reconstruction error.}
\label{fig:loss_fit}
\end{figure}

We observe that in joint weight and activation quantization, quantization errors are not merely additive. Instead, they are continuously coupled and accumulate during layer-wise propagation. Because activations exhibit a long-tail distribution, simultaneous quantization-aware training of weights and activations causes their errors to compound. This generates unstable gradients during backpropagation, disrupting binarized weight optimization and triggering model collapse. Consequently, decoupling the training into a sequential process is both theoretically essential and experimentally validated. As shown in Figure~\ref{fig:loss_fit}, there is a significant gap between quantization only quantization and joint weight \& activation quantization. So we decouple the weight quantization and activation quantization to accomplish our goal.

\section{Resource Requirements Analysis}

\begin{table}[h!]
\begin{small}
\begin{center}
\begin{tabular}{l:ccc}
\toprule
\textbf{Method} & \textbf{Tokens} & \textbf{GPUs} & \textbf{GPU hours} \\
\midrule
\textbf{FBI-LLM} & 108.5B & 32 & 22599h \\
\textbf{LBLLM} & 0.016B & 1 & 20h \\
\bottomrule
\end{tabular}
\caption{Comparison of computational resource consumption between LBLLM and binarized QAT methods.}
\label{tab:training_cost}
\end{center}
\end{small}
\end{table}

\paragraph{Training Dataset Size}
QAT-based methods such as BitNet and FBI-LLM argue that binarization leads to significant loss of pretrained knowledge in LLMs. As a result, their approach is to train a binarized LLM from scratch using large-scale corpora. In contrast, LBLLM demonstrates that a combination of PTQ and lightweight QAT can produce a binarized LLM with comparable performance. Table~\ref{tab:training_cost} compares the data requirements of QAT and our method. As shown, LBLLM requires significantly fewer training resources than QAT methods. In the supplementary material, we compare the performance of LBLLM with several QAT-based quantization methods, showing that LBLLM achieves comparable results across various metrics.


\begin{table}[h!]
\begin{small}
\begin{center}
\begin{tabular}{l:cc}
\toprule
\textbf{Model} & \textbf{FP16} & \textbf{LBLLM}\\
\midrule
\textbf{LLAMA-7B} & 13.5GB & 1.8GB \\
\textbf{LLAMA-13B} & 24.2GB & 3.2GB \\
\textbf{LLAMA-30B} & 60.5GB & 8.0GB \\
\textbf{LLAMA-65B} & 121.0GB & 16.1GB \\
\bottomrule
\end{tabular}
\caption{Inference memory compression of LBLLM on models with different parameter sizes.}
\label{tab:memory_cost}
\end{center}
\end{small}
\end{table}

\paragraph{Inference Memory Requirement}

LBLLM adopts a minimal quantization structure, without relying on additional channels to compensate for quantization errors. In Table~\ref{tab:memory_cost}, we report the inference-time memory usage of the quantized models. Compared to full-precision models, LBLLM achieves an average compression ratio of over 7×, maintaining the lowest memory footprint among similar methods.

\begin{table}[h!]
\begin{small}
\begin{center}
\begin{tabular}{l:ccc}
\toprule
\textbf{Weight Shape} & \textbf{LBLLM} & \textbf{INT8} & \textbf{INT4}\\
\midrule
\textbf{4096×4096} & 14ms & 43ms & 80ms \\
\textbf{11008×4096} & 27ms & 83ms & 158ms \\
\textbf{4096×11008} & 28ms & 101ms & 197ms \\
\bottomrule
\end{tabular}
\caption{Martix multiplication Speedup between LBLLM (W2A4) and the INT4, INT8 kernels of CUTLASS.}
\label{tab:matrix_mul_cost}
\end{center}
\end{small}
\end{table}

\paragraph{Inference Speedup}

LBLLM maintains a weight quantization configuration consistent with BWA and ARB-LLM, allowing it to leverage established CUDA kernels for inference acceleration. Although we introduce knee points for activations, this is implemented via vectorized masking with negligible computational overhead, ensuring the subsequent low-bit matrix multiplication remains identical to prior work. Table~\ref{tab:matrix_mul_cost} demonstrates the matrix computation acceleration achieved using these CUDA kernels.

\section{More Details about Our Experiment Settings}

\begin{table*}[h!]
\begin{center}
\begin{tabular}{l:ccc}
\toprule
  & \textbf{LLAMA-1-7B} & \textbf{LLAMA-2-7B} & \textbf{LLAMA-3-8B} \\
\midrule
WAR Size & 8192 & 8192 & 2048 \\
WAR Epoch & 2 & 2 & 5 \\
AAR Size & 8192 & 8192 & 8192 \\
AAR Epoch & 1 & 1 & 1 \\
W Learning Rate & 2e-5 & 2e-5 & 2e-5 \\
$\textbf{G}$ Learning Rate & 1e-4 & 1e-4 & 1e-4 \\
$\alpha, \mu$ Learning Rate & 1e-4/1e-5 & 1e-4/1e-5 & 1e-4/1e-5 \\
Clipping Ratio Learning Rate & 1e-4 & 1e-4 & 1e-4 \\
Knee Point Learning Rate & 5e-4 & 5e-4 & 5e-4 \\
GPUs & 1 & 1 & 1 \\
GPU Hours & 22 & 22 & 17 \\
\bottomrule
\end{tabular}
\caption{The configuration and training details for LBLLM. Since the learning rates for $\alpha$ and $\mu$ differ between the WAR and AAR stages, we report both values in the table.}
\label{tab:hyperparameters}
\end{center}
\end{table*}

\begin{table*}[h]
\renewcommand{\arraystretch}{1.3}
\begin{small}
\centering
\begin{tabularx}{\textwidth}{l:l:YYYYY}
\Xhline{1.2pt}
\textbf{Bits} & \textbf{Method} & \textbf{STEM}$\uparrow$ & \textbf{Humanities}$\uparrow$ & \textbf{Social Science}$\uparrow$ & \textbf{Others}$\uparrow$ & \textbf{Average}$\uparrow$ \\
\hline
FP16 & - & 36.1 & 43.3 & 51.6 & 51.8 & 45.5 \\
\hline
W(1+1)A4 & Atom & 25.9 & 24.9 & 24.0 & 26.3 & 25.3 \\
W(1+1)A4 & ARB-LLM & 30.2 & 25.4 & 30.5 & 26.0 & 27.7 \\
\hline
W(1+1)A4 & LBLLM & \textbf{26.4} & \textbf{28.8} & \textbf{27.7} & \textbf{32.0} & \textbf{28.8 }\\
\Xhline{1.2pt}
\end{tabularx}
\caption{MMLU results (\%) under the W2A4 settings on LLAMA-1-7B. Our quantization performance highlighted in bold.}
\label{tab:mmlu}
\end{small}
\end{table*}

\begin{table}[h!]
\renewcommand{\arraystretch}{1.3}
\begin{small}
\centering
\begin{tabularx}{\columnwidth}{l:l:Y}
\Xhline{1.2pt}
\textbf{Model} & \textbf{Bits} & \textbf{GSM8K} \\
\hline
\multirow{3}{*}{\textbf{LLAMA-2-7B-Chat}}
            & FP16 & 38.59 \\
            \cline{2-3}
            & BWA w/o C & 1.57 \\
            \cline{2-3}
            & LBLLM & 23.43 \\
\Xhline{1.2pt}
\end{tabularx}
\caption{GSM8K results (\%) under the W2A4 settings on LLAMA-2-7B-Chat.}
\label{tab:gsm8k_result}
\end{small}
\end{table}


This section provides additional experimental details required by the checklist but not covered in the main text. It includes information on datasets, evaluation metrics, hyperparameter settings, and computational setup used in our quantization experiments, aiming to support the reproducibility of this work. We employed fixed same random seeds for reproducibility. The numerical results from multiple runs remained consistent in this case. We did not conduct a specific random selection of numbers and the results show a clear performance gain over the baseline, which proved to be robust of our method. More detailed statistical tests will be provided in the subsequent version.

\subsection{Evaluation Metrics}

\paragraph{Memory Usage of The Quantized Model}
This section explains how we compute the memory usage of the quantized model. Given the mainstream transformer architecture used in current LLMs, our work quantizes the linear layers—which account for over 90\% of the total computation—while keeping components with relatively minor computational cost, such as embedding and activation layers, in full precision. This setting is consistent across all existing works on large language model quantization. In our memory usage calculation, we only consider the quantized linear layers. The final memory compression ratio is as follows

\begin{equation}
    \begin{aligned}
        \text{M}_q =\frac{\text{Bits}_{q}+\text{Bits}_g+\text{Bits}_p}{\text{Bits}_{fp}} \times \text{M}_{fp}
    \end{aligned}
\end{equation}

Here, $m_q$ and $m_{fp}$ denote the storage sizes of the quantized and full-precision models, respectively. $\text{Bits}_q$, $\text{Bits}_g$, $\text{Bits}_p$, and $\text{Bits}_{fp}$ represent the effective bit-widths of quantized weights, group-wise bitmaps, quantization parameters, and full-precision weights. For example, the calculation of $\text{Bits}_p$ includes the combined effective bit-width of the scaling factor $\alpha$ and the zero point $\mu$, where each group shares one $\alpha$ in 16-bit format and one $\mu$ in 1-bit format. Therefore, while the group size is 128, $\text{Bits}_p = \frac{16+1}{\text{group size}}\approx 0.148$.

\paragraph{Perplexity}
Perplexity measures how well a probability distribution or model predicts a sample. In language modeling, it reflects how closely the predicted next token matches the ground truth—the more accurate the prediction, the lower the perplexity.

\paragraph{Commonsense QA}
CommonsenseQA is a multiple-choice question answering dataset that requires diverse types of commonsense knowledge to identify the correct answers. We evaluate on PIQA, ARC, BoolQ, HellaSwag, and WinoGrande, which cover physical commonsense reasoning, middle school-level science questions, natural language inference, and general commonsense reasoning. All evaluations are conducted in a zero-shot setting.

\begin{table*}[h!]
\renewcommand{\arraystretch}{1.3}
\begin{small}
\centering
\begin{tabularx}{\textwidth}{l:l:YYY:YYYYYYY}
\Xhline{1.2pt}
\textbf{Method} & \textbf{Bits} & \textbf{Wiki} & \textbf{PTB} & \textbf{C4} & \textbf{PIQA} & \textbf{ARCe} & \textbf{ARCc} & \textbf{BoolQ} & \textbf{Hella.} & \textbf{Wino.} & \textbf{Avg.} \\
\hline
FP16 & -- & 5.47 & 22.51 & 6.97 & 76.93 & 53.58 & 40.53 & 71.07 & 72.96 & 67.17 & 63.71 \\
\hline
Onebit & W1A16 & 10.19 & -- & 11.40 & 68.01 & 42.47 & 30.20 & 57.28 & 51.54 & 58.48 & 51.33 \\
FBI-LLM & W1A16 & 9.10 & 29.60 & 10.50 & 72.60 & 53.00 & 29.90 & 61.50 & 57.70 & 58.90 & 55.60 \\
\hline
\textbf{LBLLM} & W(1+1)A4 & \textbf{9.18} & \textbf{36.79} & \textbf{10.91} & \textbf{70.97} & \textbf{44.65} & \textbf{32.51} & \textbf{64.50} & \textbf{58.54} & \textbf{58.27} & \textbf{54.91} \\
\Xhline{1.2pt}
\end{tabularx}
\caption{Perplexity and zero-shot accuracy results of different quantization methods on LLAMA-2-7B.}
\label{tab:resultwithqat}
\end{small}
\end{table*}

\begin{table*}[h!]
\renewcommand{\arraystretch}{1.3}
\begin{small}
\centering
\begin{tabularx}{\textwidth}{l:l:YYY}
\Xhline{1.2pt}
\textbf{Model} & \textbf{Bits} & \textbf{Wiki} & \textbf{PTB} & \textbf{C4} \\
\hline
\multirow{3}{*}{\textbf{Qwen-2.5-3B-Instruct}}
            & FP16 & 8.56 & 15.83 & 12.03 \\
            \cline{2-5}
            & BWA w/o C & 49206.34 & 41496.75 & 30985.61 \\
            \cline{2-5}
            & LBLLM & 18.53 & 32.97 & 25.68 \\
\Xhline{1.2pt}
\end{tabularx}
\caption{Quantized perplexity results of Qwen-2.5-3B-Instruct without KV cache quantization.}
\label{tab:qwen_result}
\end{small}
\end{table*}

\begin{table*}[h!]
\renewcommand{\arraystretch}{1.3}
\begin{small}
\centering
\begin{tabularx}{\textwidth}{l:l:YYYYYYYYYY}
\Xhline{1.2pt}
\textbf{Model} & \textbf{Bits} & \textbf{Wiki} & \textbf{PTB} & \textbf{C4} & \textbf{PIQA} & \textbf{ARCe} & \textbf{ARCc} & \textbf{BoolQ} & \textbf{Hella.} & \textbf{Wino.} & \textbf{Avg} \\
\hline
\multirow{2}{*}{\textbf{LLAMA-1-7B}}
            & FP16 & 5.68 & 27.34 & 7.08 & 77.37 & 52.48 & 41.38 & 73.06 & 73.00 & 67.01 & 64.05\\
            \cline{2-12}
            & W(1+1)A4  & 9.08 & 37.46 & 10.42 & 71.33 & 44.74 & 33.53 & 66.24 & 58.29 & 61.72  & 55.97 \\
\hline
\multirow{2}{*}{\textbf{LLAMA-1-13B}}
            & FP16 & 5.09 & 19.23 & 6.61 & 79.05 & 59.89 & 44.71 & 68.47 & 76.23 & 70.24 & 66.43\\
            \cline{2-12}
            & W(1+1)A4   & 7.90 & 38.19 & 9.23 & 74.05 & 50.21 & 34.90 & 65.81 & 64.49 & 63.85 & 58.88 \\
\Xhline{1.2pt}
\end{tabularx}
\caption{Quantized perplexity and zero-shot accuracy results of LBLLM on larger LLAMA models.}
\label{tab:diff_size_result}
\end{small}
\end{table*}

\paragraph{Massive Multitask Language Understanding}
MMLU is a challenging benchmark designed to evaluate the knowledge acquired during pretraining by testing models exclusively in zero-shot and few-shot settings. It covers 57 subjects across diverse domains, including STEM, humanities, and social sciences, with question difficulty ranging from high school to advanced professional levels. The benchmark assesses both factual knowledge and problem-solving ability, spanning topics from traditional areas like mathematics and history to specialized fields such as law and ethics. The breadth and granularity of the tasks make MMLU an ideal tool for identifying blind spots in language models.

\subsection{Hyperparameters}

In Table~\ref{tab:hyperparameters}, we present the training hyperparameter settings and more training details. Due to time constraints, we used only 2,048 samples for training LLAMA-3-8B, and most hyperparameters were chosen based on empirical heuristics rather than thorough tuning. 
We utilized the exact same hyperparameter configuration to successfully quantize models across different architectures and scales (LLaMA-1-7B, LLaMA-2-7B, LLaMA-3 8B, and LLaMA-1-13B), consistently achieving excellent results. This actually demonstrates the robustness of our quantization training framework, as its effectiveness does not rely on extensive hyperparameter tuning.
We leave comprehensive hyperparameter optimization to future work.

\subsection{Computational Setup}

Most of our experiments were conducted on NVIDIA H100 80GB with CUDA version 12.8; more detailed environment configurations will be provided in the upcoming open-source repository.

\section{Additional Experimental Results}

\subsection{Language Understading Tasks}
Table~\ref{tab:mmlu} presents a comparison of different quantization methods on the MMLU dataset. The results show that our method achieves superior overall language understanding performance and yields results closer to the full-precision model.

\subsection{Instruction-tuned benchmarks}
We added evaluation on the GSM8K benchmark with LLaMA-2-7B-Chat using LBLLM, to demonstrate the capability on complex reasoning tasks (Table~\ref{tab:gsm8k_result}). The quantized model achieves a score of 37.42\%, significantly outperforming BWA in accuracy under identical settings with no extra high-bit channels. This indicates that our W(1+1)A4 quantization effectively preserves the reasoning capabilities required for instruction-following tasks. We will provide additional experimental results and compare them with relevant baselines.

\subsection{Compared with QAT Method}

We compare our results with two binarized QAT methods, Onebit and FBI-LLM; due to the high computational requirements of these approaches, we directly adopt the settings and results reported in their papers. Across various metrics, our method outperforms binarized QAT approaches in many aspects and achieves comparable performance on the remaining ones, demonstrating its strong potential.

\subsection{Evaluation breadth across different model families} 

To demonstrate that our method is inherently architecture-agnostic, we included experimental results for Qwen-2.5-3B (Table~\ref{tab:qwen_result}), proving that LBLLM generalizes seamlessly beyond the LLaMA family. By focusing on standard Transformer linear layers, LBLLM bypasses the need for architecture-specific workarounds, such as mixed-precision extra channels or rotational matrices. For comparison, we present the results of 'BWA w/o C', which represents the quantization performance of the baseline method BWA under identical settings, but without the additional tricks.

\subsection{Larger Model Results}

We further evaluated LBLLM on LLaMA models with varying parameter sizes. The results in Table~\ref{tab:diff_size_result} demonstrate that even on relatively larger models (e.g., LLaMA-13B), LBLLM achieves robust quantization performance through lightweight training. Furthermore, LBLLM exhibits a narrower performance gap compared to the full-precision baseline on these larger models. Due to computational resource constraints, evaluations on larger-scale models are reserved for future work.

\end{document}